\documentclass{article}

\usepackage{microtype}
\usepackage{graphicx}
\usepackage{booktabs}
\usepackage{amsmath}
\usepackage{acronym}

\DeclareMathOperator{\sign}{sgn}

\usepackage{hyperref}

\usepackage[accepted]{mlsys2023}

\mlsystitlerunning{Improved in-memory training}

\newcommand{\algref}[1]{Alg.~\ref{alg:#1}}
\newcommand{\figref}[1]{Fig.~\ref{fig:#1}}
\newcommand{\secref}[1]{Sec.~\ref{sec:#1}}
\renewcommand{\eqref}[1]{Eq.~\ref{eq:#1}}
\newcommand{\lmax}{l_\text{max}}
\newcommand{\weightgran}{\delta_w}
\newcommand{\every}{n_s}
\newcommand{\bmax}{b_\text{max}}
\newcommand{\bmin}{b_\text{min}}
\newcommand{\bdtod}{\sigma_b}
\newcommand{\transfercount}{t}
\newcommand{\autogran}{\gamma_0}
\newcommand{\leakylambda}{\beta}
\newcommand{\inchopprob}{\rho}
\newcommand{\rerammodel}[2]{dF\left(#1| \boldsymbol{\theta}, \,{}#2\right)}
\newcommand{\analog}[1]{\breve{#1}}
\newcommand{\dwmin}{\weightgran}
\newcommand{\updowndtod}{\sigma_\pm}
\newcommand{\dwmindtod}{\sigma_\text{d-to-d}}
\newcommand{\dwminstd}{\sigma_\text{c-to-c}}
\newcommand{\sym}[1]{\analog{#1}^*}
\newcommand{\weighterror}{\epsilon_w}
\newcommand{\nstates}{n_\text{states}}

\acrodef{TTII}[TTv2]{Tiki-Taka version 2}
\acrodef{TTIII}[c-TTv2]{Chopped-TTv2}
\acrodef{TTIV}[AGAD]{Analog Gradient Accumulation with Dynamic reference}
\newcommand{\TTii}{\ac{TTII}}
\newcommand{\TTiifull}{\acf{TTII}}
\newcommand{\TTiii}{\ac{TTIII}}
\newcommand{\TTiiifull}{\acf{TTIII}}
\newcommand{\TTiv}{\ac{TTIV}}
\newcommand{\TTivfull}{\acf{TTIV}}
\acrodef{AIHWKIT}[\textsc{AIHWKit}]{IBM Analog Hardware Acceleration Kit}
\acrodef{FP}[FP]{floating point}
\acrodef{SGD}[SGD]{stochastic gradient descent}
\acrodef{DNN}[DNN]{deep neural network}
\acrodef{MVM}[MVM]{matrix-vector multiplication}
\acrodef{SP}[SP]{symmetry point}
\acrodef{NVM}[NVM]{non-volatile memory}
\acrodef{ADC}[ADC]{analog-to-digital converter}

\newcommand{\AIHWKit}{\ac{AIHWKIT}}
\newcommand{\pytorch}{\mbox{\textsc{PyTorch}}}

\begin{document}

\twocolumn[
\mlsystitle{Fast offset corrected in-memory training}
\mlsyssetsymbol{equal}{*}

\begin{mlsysauthorlist}
\mlsysauthor{Malte J. Rasch}{ibm}
\mlsysauthor{Fabio Carta}{ibm}
\mlsysauthor{Omebayode Fagbohungbe}{ibm}
\mlsysauthor{Tayfun Gokmen}{ibm}
\end{mlsysauthorlist}

\mlsysaffiliation{ibm}{IBM Research, Yorktown Heights, New York, USA.}

\mlsyscorrespondingauthor{Malte J. Rasch}{malte.rasch@ibm.com}
\mlsyscorrespondingauthor{Tayfun Gokmen}{tgokmen@us.ibm.com}

\mlsyskeywords{Machine Learning, In-memory computing, Analog AI}

\vskip 0.3in

\begin{abstract}
  In-memory computing with resistive crossbar arrays has been
  suggested to accelerate deep-learning workloads in highly efficient
  manner. To unleash the full potential of in-memory computing, it is
  desirable to accelerate the training as well as inference for large
  \acp{DNN}. In the past, specialized in-memory training algorithms
  have been proposed that not only accelerate the forward and backward
  passes, but also establish tricks to update the weight in-memory and
  in parallel. However, the state-of-the-art algorithm (\TTii) still
  requires near perfect offset correction and suffers from potential
  biases that might occur due to programming and estimation
  inaccuracies, as well as longer-term instabilities of the device
  materials. Here we propose and describe two new and improved
  algorithms for in-memory computing (\TTiii\ and \TTiv), that retain
  the same runtime complexity but correct for any remaining offsets
  using choppers. These algorithms greatly relax the device
  requirements and thus expanding the scope of possible materials
  potentially employed for such fast in-memory \ac{DNN} training.
\end{abstract}
]

\printAffiliationsAndNotice{}  
\section{Introduction}
\label{introduction}

In-memory computing with resistive crossbar arrays could greatly
accelerate deep-learning workloads, since energy efficient
acceleration can be achieved by implementing ubiquitous \acp{MVM}
using resistive elements and fundamental physics (Kirchhoff's and
Ohm's laws)~\cite{Y2020sebastianNatNano, burr2017neuromorphic,
  haensch2019next, yang2013memristive, sze2017efficient}. While most
prototype chip building efforts to-date have been focused on
accelerating the inference phase of pre-trained
\acp{DNN}~\cite{wan2022compute, khaddam2021hermes, xue2021cmos,
  fick2022analog, narayanan2021fully, ambrogio2018equivalent,
  yao2020fully}, in terms of raw compute requirements, the training
phase typically is orders of magnitude more expensive than inference,
and thus would in principle have a greater need for hardware
acceleration using in-memory compute~\cite{gokmen2016}. However,
in-memory training using non-volatile memory elements has been
challenging, in particular because of the asymmetric and non-ideal
switching of the memory devices, and the much greater precision
requirements during gradient update (see e.g. \cite{jain2019neural}
for a discussion).

\begin{figure*}[t]
  \includegraphics[width=\textwidth]{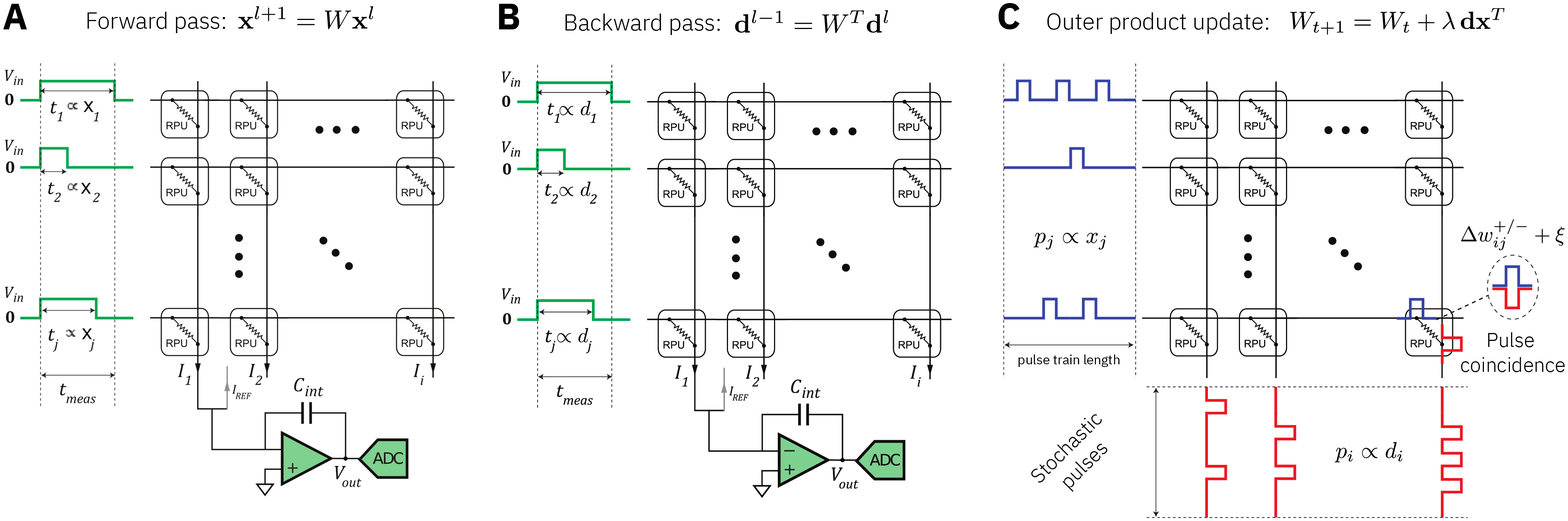}
  \vspace{-0.3cm}
  \caption{Physical implementaton of in-memory \acf{MVM} for forward and backward passes and
    outer product update according to~\cite{gokmen2016}. }
  \label{fig:fwd_bwd_upd}
\end{figure*}

To accelerate the \ac{SGD} of \ac{DNN} training, not only the forward pass, but also the
backward pass, as well as gradient computation and update needs to be
considered. While the backward pass of a linear layer is
straightforwardly accelerated in-memory by transposing the inputs and
outputs, the gradient accumulation and update is more difficult to
perform in a fast and efficient manner in-memory. One way is to
sacrifice speed and efficiency by computing the gradient and its
accumulation in digital and only accelerating
the forward and backward pass in-memory, as suggested
by~\cite{nandakumar2018, nandakumar2020}. On the other hand, Gokmen et
al.~\cite{gokmen2016} suggested to use the coincidence of stochastic
pulses to efficiently perform the outer product update operation
fully in-memory in a highly efficient and parallel manner. However, in
its initial naive formulation, a bi-directionally switching device of
high symmetry and precision (in terms of reliability of incremental
conductance changes) is needed to be able to compute and
accumulate the gradients with the required
accuracy~\cite{agarwal2016resistive, gokmen2016}.  Further refinements
of the algorithm, however, relaxed the
requirements of the number of reliable conductance state and the
device symmetry considerably~\cite{gokmen2020, gokmen2021}.

In the latest installment, the so-called \TTii\ learning
algorithm~\cite{gokmen2021}, three tunable conductance elements for
each weight matrix element are required, namely the
matrices\footnote{We write $\analog{X}$ for a weight matrix $X$ that
  is thought of coded into the conductances of a crossbar array, to
  distinguish between matrices that are in digital memory.}
$\analog{A}$, $\analog{R}$, and $\analog{W}$. The first two
conductances, $\analog{A}$ and $\analog{R}$, are used to perform the
outer product and (parts of) the gradient accumulation, whereas
$\analog{W}$ is used as the representation of the weight $W$ of a
linear layer and thus used in the forward and backward passes. On
functional level, the algorithm is similar to modern \ac{SGD} methods
that introduce a momentum term (such as ADAM~\cite{kingma2014adam}),
since also here the gradient is first computed and accumulated in a
leaky fashion onto a separate matrix before being added to the weight
matrix. However, the analog friendly \TTii\ algorithm computes and
transfers the accumulated gradients asynchronously for each row (or
column) to gain run-time advantages. Furthermore, crucially, the
device asymmetry of the memory element causes an {\it input-dependent}
decay of the recent accumulated gradients as opposed to the usual
constant decay rate of the momentum term that is difficult to
efficiently implement in-memory~(see also discussion
in~\cite{gokmen2021, onen2022neural}).

In more detail, for each update, the outer product gradient update is
first computed onto $\analog{A}$ using coincidences of stochastic
pulse trains. Then $\analog{A}$ is read in a row-wise (or column-wise)
fashion and the read-out accumulated gradients are additionally
filtered onto a digital storage (hidden) matrix $H$ to smooth high
frequency noise introduced by the gradient computation itself as well
as the physical writing process using stochastic pulses on noisy
materials. Only when a threshold of the accumulated gradient is
reached, a single writing pulse is sent to the corresponding weight
element of the matrix $\analog{W}$.

While this \TTii\ algorithm greatly improves the material
specifications by introducing low-pass filtering of the recent
gradients, it hinges on the assumption that the device has a
pre-defined and stable \ac{SP} within its conductance
range~\cite{onen2022neural}. The \ac{SP} is defined as the conductance
value where a positive and a negative update will result on average in
the same net change of the conductance. Because of the assumed device
asymmetry, the SP acts as a stable fix point for random inputs, which
causes the accumulated gradient on $\analog{A}$ to automatically decay
near convergence. However, to induce a decay towards zero
algorithmically, it is essential to identify the SP with the zero
value for each device, which is achieved by removing the offset using a
reference array $\analog{R}$.  The reference conductance $\analog{R}$
is thus used to store the SP values of its corresponding devices of
$\analog{A}$ and when accessing $\analog{A}$, instead of directly
reading $\analog{A}$, the difference $\analog{A} - \analog{R}$ is read
instead, while only $\analog{A}$ is updated during training.

Programming the offset array $\analog{R}$ to the SP of $\analog{A}$
once at the beginning could be problematic in practice since any
remaining offset $\analog{A} - \analog{R} + \Delta$ would be
continuously read and added onto the weight matrix $\analog{W}$. Such
constant offset $\Delta$ from the fixed point might arise for instance
by an inaccurate estimation or writing of the SP onto $\analog{R}$ or
a long-term temporal variation the SP of $\analog{A}$.  Moreover,
offset transients caused by temporally $\analog{A} \neq \analog{R}$
might result in significant gradient accumulation and writing onto
$\analog{W}$, thus possibly leading to oscillations and non-optimal
learning behavior.

In this paper, we propose two new algorithms based on \TTii, that
improve the gradient computation in case of any kinds of reference
instability or residual offsets. Both algorithms introduce a technique
borrowed from amplifier circuit design, called
chopper~\cite{enz1996circuit}. A chopper is a well-known technique to
remove any offsets or residuals caused by the accumulating system that
are not present in the signal, by modulating the signal with a random
sign-change (the chopper) that is then corrected for when reading from
the accumulator.  In our first algorithm, we randomly flip the sign of
the activation or error vector before computing the outer product
update in-memory. After reading $\analog{A} - \analog{R}$, the read
value is then again sign-corrected. In this way, any remaining offset
residuals are averaged out by the filtering stage without
significantly increasing the complexity of the algorithm. In the
second new algorithm, we propose to additionally use an on-the-fly
dynamic estimation of the recent reference point that further improves
on removing transients and essentially obliterates the requirement of
having a tunable reference array $\analog{R}$ altogether.
\begin{algorithm}[tb]
  \caption{Parallel in-memory update using dynamically adjusted stochastic
    pulsing.}
  \label{alg:stocgradient}
  \label{alg:sgd}

   \begin{algorithmic}
     \renewcommand{\SET}{\STATE {\bf set} }
     \newcommand{\GOTO}{\STATE {\bf goto} }
     \INPUT{data vector $x_i$, size $n$, gradient vector $d_j$, size $m$, weight
       matrix $\analog{w}_{ij}$}, average pulse size at SP $\weightgran$, learning rate
     $\eta$, max pulses $\lmax$
     \OUTPUT{Updated analog weight matrix $\analog{w}_{ij}$}
     \SET $m_x \leftarrow \max_i |x_i|$ and $m_d \leftarrow \max_j |d_j|$
     \SET $\kappa \leftarrow \frac{\eta  m_x m_d}{\weightgran }$
     \SET $l \leftarrow \min(\lmax,
     \left\lceil\kappa\right\rceil)$ and $\tilde{m}_d\leftarrow m_d \min(\frac{\lmax}{\kappa}, 1)$
     \SET $a \leftarrow \sqrt{\frac{\eta \,m_x}{l\,\tilde{m}_d
         \weightgran}}$ and $b \leftarrow \sqrt{\frac{\eta \,\tilde{m}_d}{\lmax \,m_x \weightgran}}$
     \FOR{$k=1$ {\bf to} $l$}
     \SET all $q_i \leftarrow 0$ and all $p_j \leftarrow 0$
     \FOR{$i=1$ {\bf to} $m$}
     \STATE Draw random number $\xi \in {\cal U}(0, 1)$
     \IF{$a |d_i| < \xi$}
     \SET $q_i\leftarrow \sign(d_i)$
     \ENDIF
     \ENDFOR
     \FOR{$j=1$ {\bf to} $n$}
     \STATE Draw random number $\xi \in {\cal U}(0, 1)$
     \IF{$b |x_j| < \xi$}
     \SET $p_j\leftarrow\sign(x_j)$
     \ENDIF
     \ENDFOR

     \STATE \COMMENT{Send pulse vectors $\mathbf{p}$ and $\mathbf{q}$ to
     update $\analog{w}_{ij}$:}
     \FOR{$i=1$ {\bf to} $m$}
     \FOR{$j=1$ {\bf to} $n$}
     \IF{$p_j\neq 0$ and $q_i \neq 0$}
     \SET $\analog{w}_{ij}\leftarrow \analog{w}_{ij} + \rerammodel{\analog{w}_{ij}}{\sign(q_ip_j)}$
     \ENDIF
     \ENDFOR
     \ENDFOR
     \ENDFOR
\end{algorithmic}
\end{algorithm}

\section{Background: In-memory optimizer}
\subsection{Analog \acl{MVM}}

Using resistive crossbar arrays to compute an \ac{MVM} in-memory has
been suggested early on~\cite{steinbuch1961lernmatrix}, and multiple
prototype chips where \acp{MVM} of \acp{DNN} during inference are
accelerated have been described~\cite{wan2022compute,
  khaddam2021hermes, xue2021cmos, fick2022analog,
  narayanan2021fully}. In principle, in all these solutions, the
weights of a linear layer are stored in a crossbar array of tunable
conductances, inputs are encoded e.g. in voltage pulses and Ohm's and
Kirchhoff's law are used to multiply the weights with the inputs and
accumulate the products, respectively~(\figref{fwd_bwd_upd}A, see also
e.g. \cite{Y2020sebastianNatNano} for more details). In many designs,
the resulting currents or charge, is converted back to digital by
highly parallel \acp{ADC}.

For fully in-memory analog training, as suggested
in~\cite{gokmen2016}, additionally a transposed \ac{MVM} has to be
implemented for the backward pass, which can be achieved by
transposing input and output accordingly (see \figref{fwd_bwd_upd}B).

\subsection{In-memory outer product update }
While accelerating the forward and the backward pass of \ac{SGD} using
in-memory \acp{MVM} is promising, for a full in-memory training
solution, in-memory gradient computation and weight update have to be
considered for acceleration as well: it is the remaining operation
having ${\cal O}(n^2)$ complexity (where $n \times n$ is the size of
the matrix in ``analog'') during \ac{DNN} training~\cite{gokmen2016}.

For the gradient accumulation of a weight matrix $W$ of a linear layer
(i.e. computing $\mathbf{y} = W\mathbf{x}$), the outer product update
of $W \leftarrow W + \lambda\mathbf{d}\mathbf{x}^T$
needs to be computed. While this can be done in digital, possibly
exploiting sparseness~(e.g. see \cite{nandakumar2020}), it would
still require ${\cal O}(n^2)$ digital operations, and doing so would
thus limit the overall acceleration factor obtainable for in-memory
training. To accelerate also the outer-product update to be performed
in-memory and fully parallel, Gokmen et al.~\cite{gokmen2016}
suggested to use stochastic pulse trains and their coincidence (as
illustrated in \figref{fwd_bwd_upd}~C).

The exact update algorithm has gone through a number of improvements
in recent years~\cite{gokmen2017cnn, gokmen2018lstm}, however, we here
suggest to use a yet improved version in
\algref{stocgradient}. In particular, the pulse trains are dynamically
adjusted in length in our version. Note that we here assume for
simplicity that negative pulses can be sent. In practice negative and
positive pulses are sent in sequential phases.

While \cite{gokmen2016, gokmen2017cnn, gokmen2018lstm} have
investigated the noise properties when using \algref{stocgradient} to
directly implement the gradient update in-memory, it turns out that
this would require very symmetric switching characteristics of the
memory device elements. Thus, the requirements of such a ``in-memory
plain \ac{SGD}'' algorithm turns out to be challenging in particular
in face of the asymmetry observed in today's materials, which we
discuss in the next section.

\subsection{Device material model }

When subject to a large enough voltage pulse, bi-directionally
switching device materials, such as resistive random access memories
(ReRAM)~\cite{zahoor2020resistive}, typically show a high degree of
asymmetry in one versus the other direction and gradual saturation to
a minimal maximal conductance value. In previous
studies~\cite{fusi2007limits, frascaroli2018evidence}, it was shown
that the ``soft-bounds'' model characterizes the switching behaviour
qualitatively well. According to that model, the weight change
$w \leftarrow w + \rerammodel{w}{D}$ to a single update pulse in
either up ($D=+$ or down $D=-$) direction is given by
\begin{equation}
  \label{eq:softbounds}
  \begin{aligned}
   \rerammodel{w}{+} &=& \alpha_{+}\frac{\bmax - w}{\bmax} (1 +
  \dwminstd \xi)\\
  \rerammodel{w}{-} &=& {} - \alpha_{-}\frac{\bmin - w}{\bmin} (1 +
  \dwminstd \xi)
\end{aligned}
\end{equation}
Here we use the placeholder $\boldsymbol{\theta}$ for the
hyper-parameters.  To capture device-to-device variability, we
initially draw $\bmax = \max(1 + \bdtod\xi_1, 0)$ and
$\bmin = \min(-1 + \bdtod\xi_2, 0)$ where the
$\xi_i \in {\cal N}(0, 1)$ are random numbers that are different for
each device. The slope parameters are given by
\begin{equation}
  \label{eq:slope}
  \begin{aligned}
  \alpha_+ &=& \dwmin \,\left(\gamma + \rho\right)\\
  \alpha_- &=& \dwmin \, \left(\gamma - \rho\right)
  \end{aligned}
\end{equation}
where $\gamma = e^{\dwmindtod\xi_3}$, and $\rho=\updowndtod \xi_4$, so
that $\dwmindtod$ is a hyper-parameter for the variation of the slope
across devices, and $\updowndtod$ a separate device-to-device
variation in the difference of the slope between up and down
direction.  The material parameter $\dwmin$ determines the average
update response for one pulse when the weight is at zero. We define
the number of device states by the average weight range divided by
$\dwmin$, ie. $\nstates = \frac{2}{\dwmin}$.

\subsubsection{Symmetry point}
It can easily be seen that for the device model \eqref{softbounds},
the update size linearly decreases for positive updates up the bound
$\bmax$, and likewise linearly decreases for negative updates down to
the bound $\bmin$, and thus the weight will saturate at $\bmin$ and
$\bmax$.  Thus there exists a weight value at which the up and down
updates sizes are equal on average, which is thus called the
\acf{SP}~\cite{gokmen2020, onen2022neural}.

Solving \eqref{softbounds} for the SP $\sym{w}$ by requiring up and
down pulse sizes to be equal, one finds\footnote{we assume here only
  the non-degenerated case, i.e. $\bmax > 0$, $\bmin < 0$,
  $\alpha_{+}>0$, and $\alpha_{-}>0$}
\begin{equation}
  \label{eq:sympoint}
   \sym{w}= \frac{\alpha_{+} - \alpha_{-}}{\frac{\alpha_{+}}{\bmax} - \frac{\alpha_{-}}{\bmin}}
      = \frac{2 \rho}{\frac{\gamma + \rho}{\bmax} - \frac{\gamma- \rho}{\bmin}}
\end{equation}
In summary, we use a device model that is already defined in a way
that the \ac{SP} is at zero when $\updowndtod=0$. However, if
$\updowndtod > 0$ the \ac{SP} is not at zero.

\paragraph{Transients when decaying towards the \ac{SP}}
If random up-down pulsing (without a bias in one direction) is applied
to the devices, the device will reach its \ac{SP}. In the case where a
positive pulse always follows a negative pulse, the weight change
(assuming for the moment
$\sigma_b=\dwminstd=\updowndtod=\dwmindtod=0$) can be written as:
\begin{eqnarray}
  \label{eq:spdecay}
   \Delta w &=& \rerammodel{w}{+} + \rerammodel{w + \rerammodel{w}{+}}{-} \nonumber\\
            &\approx& \rerammodel{w}{+} + \rerammodel{w}{-} \nonumber\\
            &=& \dwmin (1 - w)  - \dwmin (1 + w)\nonumber\\
            &=& -2 \dwmin w
\end{eqnarray}
which shows that for repeated pairs of up and down pulses the
weight will decay exponentially with (approximate) decay rate of
$\tau = 2 \dwmin$ to a fixed point at $\sym{w}=0$. 

\paragraph{Setting the reference device to the \ac{SP}}
To remove the offsets arising from the decay towards the \ac{SP}, some
algorithms discussed below require a the reference matrix $\analog{R}$
to be set to the \ac{SP}~\cite{gokmen2021, onen2022neural}. We model
this by setting (with \eqref{sympoint})
\begin{equation}
  \label{eq:reference}
   \analog{r}_{ij}= \sym{a}_{ij} + \mu_r + \sigma_r \xi_{ij}
\end{equation}
where $\xi_{ij}\in {\cal N}(0, 1)$. Thus $o_r = \mu_r + \sigma_r \xi$
models the remaining offsets on the reference device after \ac{SP}
subtraction.  Remaining offsets could be due to incorrectly estimating
the \ac{SP} of $\analog{A}$ or to the programming error when setting
$\analog{R}$. The algorithm of how to program $\analog{R}$ to the SP
of $\analog{A}$ in practice is discussed in~\cite{kim2019zero}.

\begin{algorithm}[t!]
  \caption{\TTiiifull\ algorithm.
  }
  \label{alg:TTii}
  \label{alg:TTiii}
   \begin{algorithmic}
     \renewcommand{\SET}{\STATE {\bf set} }
     \newcommand{\GOTO}{\STATE {\bf goto} }
     \newcommand{\CALL}{\STATE {\bf call} }
     \newcommand{\DRAW}{\STATE {\bf draw} }
     \INPUT{learning rate $\lambda$, buffer scale $\autogran$,
       transfer period $\every$, chopper probability $\inchopprob$}

     \OUTPUT{converged analog weights $\analog{w}_{ij}$}
     \SET $\gamma \leftarrow \autogran \frac{\weightgran}{n \every}$
     \SET counters $s$ and $k$ to 0 and digital matrix $h_{ij}\leftarrow 0$
     \CALL Program reference $\analog{R}$ to the SP of $\analog{A}$:
     $\analog{r}_{ij} \leftarrow \analog{a}_{ij}^*$
     \REPEAT

     \CALL start \ac{SGD} iteration for mini-batch \COMMENT{(using $\analog{w}_{ij}$
       as weights) until the update of $\analog{w}_{ij}$ is required}

     \FORALL{inputs $\mathbf{x}$ and $\mathbf{d}$ to be updated onto $\analog{W}$}

     \CALL \algref{stocgradient} to update $(c_jx_j)$ and $\mathbf{d}$ onto $\analog{A}$

     \SET $s \leftarrow s + 1$

     \IF{$s = \every$}

     \SET $k \leftarrow k + 1 \mod n$ and $s \leftarrow 0$ and all $p_i\leftarrow 0$

     \CALL the analog \ac{MVM}: $\mathbf{y}\leftarrow (\analog{A} - \analog{R})\, \mathbf{e}_{k}^{(n)}$

     \FOR{$i=1$ {\bfseries to} $m$}

     \SET $h_{ik}\leftarrow h_{ik} + c_k \frac{\lambda}{\gamma} y_i$

     \IF{$|h_{ik}| > 1$}
     \SET $p_i \leftarrow \sign(h_{ik})$
     \SET $h_{ik} \leftarrow 0$
     \ENDIF

     \ENDFOR

     \STATE \COMMENT{Use pulse vector $p_i$ to update column $k$ of $\analog{W}$:}

     \FOR{$i=1$ {\bfseries to} $m$}
     \IF{$p_i\neq0$}
     \SET
     $\analog{w}_{ik}\leftarrow \analog{w}_{ik} + \rerammodel{\analog{w}_{ik}}{\sign(p_i)}$
     \ENDIF
     \ENDFOR
     \STATE \COMMENT{Flip the chopper as follows.}
     \DRAW random number $\xi \in {\cal U}(0, 1)$
     \IF{$\inchopprob < \xi$}
     \SET $c_{k} \leftarrow -c_{k}$
     \ENDIF

     \ENDIF

     \ENDFOR
     \CALL finish \ac{SGD} iteration
     \UNTIL{convergence is reached}
\end{algorithmic}
\end{algorithm}

\subsection{\TTii\ algorithm}
To lower the device requirements for in-memory \ac{SGD},
\cite{gokmen2020} proposed to first compute the outer product update
in a fast manner in-memory and thus accumulate the recent past of the
gradients onto a separate analog crossbar array $\analog{A}$, but
slowly transfer the recent accumulated gradients by sequential vector
reads (single columns or rows) of $\analog{A}$ onto the analog weight
matrix $\analog{W}$, to counter-act the loss of the information due to
the device asymmetry on the gradient matrix $\analog{A}$.  The
innovative aspect of this approach is that the accumulated gradients
on $\analog{A}$ are decaying in an input dependent manner due to the
device asymmetry: as discussed above, random (unbiased) input will
cause the conductances on $\analog{A}$ to decay to the \acp{SP}. Thus,
if the \acp{SP} are subtracted, the gradient matrix decays to zero by
design when no net gradients are present as is the case at a minimum
in the loss landscape.

Furthermore, since the transfer of the gradients on $\analog{A}$ to
$\analog{W}$ is done with only ${\cal O}(n)$ digital operations (at
most one vector per matrix update) the digital and additional
computations introduced by the transfer read and write to $\analog{W}$
is not a bottleneck, as opposed to the case where the full gradient
update would be done in digital that required ${\cal O}(n^2)$
operations.

In~\cite{gokmen2021}, the algorithm was improved by additionally
filtering the transferred gradient using a digital matrix $H$ that
additionally accumulates the recently read gradients from $\analog{A}$
and, only once a threshold level is reached, a pulse is sent to update
corresponding device of the weight matrix $\analog{W}$. This
additional averaging, while introducing more digital computations,
still retains the ${\cal O}(n)$ character, as at most one vector is
operated on for each transfer. It was shown~\cite{gokmen2021} that the
filtering stage greatly lowers the device requirements, in particular
the effective number of device states: only 15 states are enough to
train an LSTM to acceptable accuracy. This relaxed device requirements
now make it possible to think about current ReRAM devices for
in-memory training~\cite{gong2022deep}.
This \TTiifull\ algorithm is our baseline comparison\footnote{Note
  that the detailed implementation of \TTii\ can be inferred from
  \algref{TTii} and \figref{algillu} by setting $\inchopprob=0$ and
  thus fixing the choppers to $c_k=1$ as discussed in
  \secref{improved-opt}.}.

\section{Improved in-memory optimizers}
\label{sec:improved-opt}
We here present our new algorithmic contributions as well as a new way
to scale the learning rates depending on material specifications and
weight matrix.

\subsection{\TTiii\  algorithm}
While an analog reference matrix $\analog{R}$ has the advantage to
subtract the \ac{SP} from $\analog{A}$ efficiently, the design choice
comes with unique challenges. In particular, the programming of
$\analog{R}$ might be inexact or the \ac{SP} might vary on a slow time
scale, introducing some remaining offset transients. However, any
offsets would constantly accumulate on $H$ and be written onto in
$\analog{W}$ thus biasing the weight matrix unwantedly. Moreover, the
decay of $\analog{A}$ to its \ac{SP} is the slower the more states the
device has (see \eqref{spdecay}) and input dependent. The latter might
cause the $\analog{A}$ to not decay for many updates which again would
cause a bias in $\analog{W}$ (as shown below). While feedback from the
loss would eventually change the gradients and correct $\analog{W}$,
the learning dynamics might nevertheless be impacted.

\begin{algorithm}[t!h]
  \caption{\TTivfull\ algorithm.
  }
   \label{alg:TTiv}
   \begin{algorithmic}
     \renewcommand{\SET}{\STATE {\bf set} }
     \newcommand{\CALL}{\STATE {\bf call} }
     \newcommand{\DRAW}{\STATE {\bf draw} }

     \INPUT{learning rate $\lambda$, buffer scale $\autogran$,
       transfer period $\every$, chopper frequency $\inchopprob$,
       leaky-average time-scale $\leakylambda$}

     \OUTPUT{converged analog weights $\analog{w}_{ij}$}
     \SET $\gamma \leftarrow \autogran \frac{\weightgran}{n \every}$
     \SET counters $s$, $k$ and $\transfercount$ to 0 and all choppers $c_j\leftarrow1$
     \SET digital matrices $h_{ij}$, $\mu_{ij}$ and $\mu^\text{past}_{ij}$ to all 0

     \REPEAT
     \CALL start \ac{SGD} iteration
     \FORALL{inputs $\mathbf{x}$ and $\mathbf{d}$ to be updated onto $\analog{W}$}

     \CALL \algref{stocgradient} to update $c_j x_j$ and $d_i$ onto $\analog{a}_{ij}$

     \SET $s \leftarrow s + 1$

     \IF{$s = \every$}
     \SET $s \leftarrow 0$ and $\transfercount \leftarrow \transfercount + 1$ and
     all $p_i \leftarrow 0$
     \SET $k \leftarrow k + 1 \mod n$
     \CALL analog \ac{MVM}: $\mathbf{y}\leftarrow\analog{A}\, \mathbf{e}_{k}$
     \FOR{$i=1$ {\bf to} $m$}
     \SET $h_{ik}\leftarrow h_{ik} + c_k\,\frac{\lambda}{\gamma} \left(y_i - \mu^\text{past}_{ik}\right)$
     \SET $\mu_{ik} \leftarrow (1 - \leakylambda)\, \mu_{ik} + \leakylambda y_i$
     \IF{$|h_{ik}| > 1$}

     \SET $p_i \leftarrow \sign(h_{ik})$
     \SET $h_{ik} \leftarrow 0$
     \ENDIF

     \ENDFOR

     \STATE \COMMENT{Use pulse vectors $p_i$ to update column $k$ of $\analog{W}$:}
     \FOR{$i=1$ {\bfseries to} $m$}
     \IF{$p_i\neq0$}
     \STATE
     $\analog{w}_{ik}\leftarrow \analog{w}_{ik} + \rerammodel{\analog{w}_{ik}}{\sign(p_i)}$
     \ENDIF
     \ENDFOR

     \STATE \COMMENT{Flip chopper and reset $\mu_{ij}$ and $\mu_{ij}^\text{past}$ as follows.}
     \IF{$\transfercount \mod \left\lceil\frac{1}{\inchopprob}\right\rceil = 0$}
     \SET $c_{k} \leftarrow -c_{k}$
     \FOR{$i=1$ {\bfseries to} $m$}
     \SET $\mu_{ik}^\text{past} \leftarrow \mu_{ik}$
     \SET $\mu_{ik} \leftarrow 0$
     \ENDFOR
     \ENDIF

     \ENDIF
     \ENDFOR
     \CALL finish \ac{SGD} iteration
     \UNTIL{convergence is reached.}
   \end{algorithmic}
 \end{algorithm}

\begin{figure*}[t]
  \includegraphics[width=\textwidth, clip, trim= 0 0.05cm 0 0]{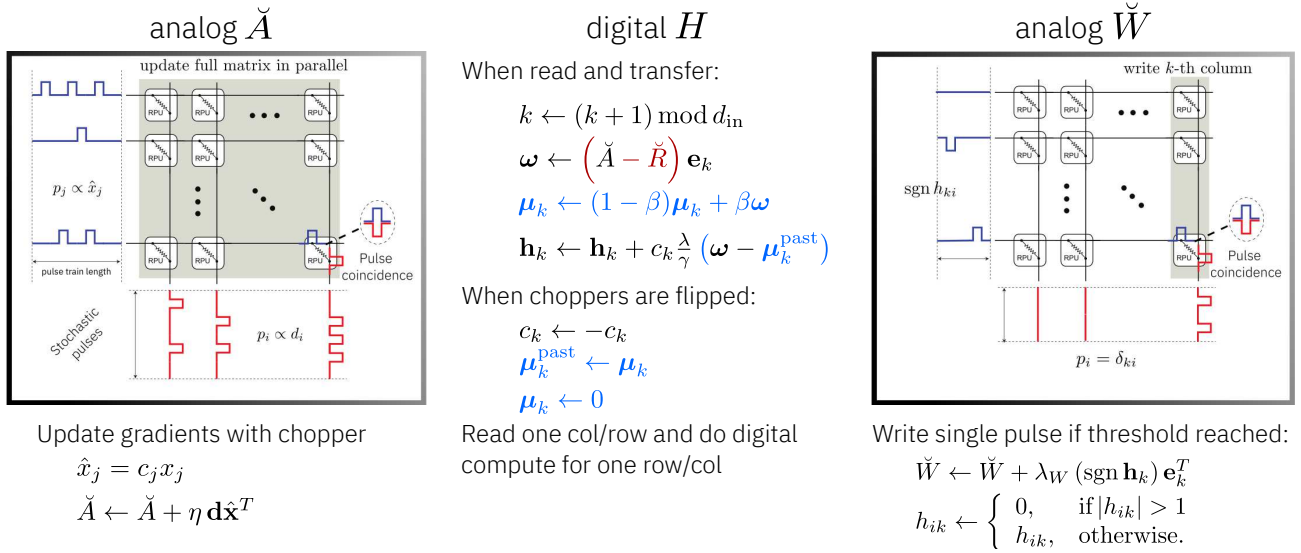}
  \vspace{-0.6cm}
  \caption{Improved in-memory training algorithms, \TTiii\ and
    \TTiv. The blue color indicates equations unique and in addition
    for \TTiv. The reference subtraction (in red) is not needed for
    \TTiv. Removing blue equations and fixing the choppers recovers
    \TTii. }
  \label{fig:algillu}
\end{figure*}

 To improve on any remaining bias and low-frequency noise and bias
 sources, we suggest here to improve the algorithm by introducing
 choppers. Chopper stabilization is a common method for offset
 correction in amplifier circuit
 design~\cite{enz1996circuit}. Choppers modulate the incoming signal
 before gradient accumulation, and subsequently demodulate during the
 reading of the accumulated gradient. In more detail, we introduce
 choppers $c_j\in\{-1, 1\}$ that flip the sign of each of the
 activations $x_j$ before the gradient accumulation on
 $\analog{A}$. When reading the $k$-th column of $\analog{A}$ to be
 transferred onto $H$ (as in \TTii), we apply the corresponding
 chopper $c_k$ to recover the correct sign of the signal. The choppers
 are then flipped randomly, see \algref{TTiii} for the detailed
 algorithm (compare also to \figref{algillu}).

In this manner, any low frequency component that is caused by the asymmetry or
any remaining offsets and transients on $\analog{A}$ is not
modulated by the chopper and thus canceled out by the sign flips. We call this
algorithm \TTiiifull\ stochastic gradient descent.

\subsection{\TTiv\  algorithm}

While the chopper together with the low-pass filtering greatly improve
the resilience to any remaining offsets (see \secref{results}), if
offsets are too large the filtering required might be too slow, thus
reducing the responsiveness of the algorithm. This can be understood as follows.
Assume that a constant gradient $g$ is calculated onto an single
element of the gradient matrix, $\analog{a}$. Assume further two
chopper phases with $k$ reads each and a reference offset
$o_r$. Then corresponding element in the hidden matrix $h$
will receive the following changes
\begin{eqnarray}
  h_{t+n} - h_{t} &=&  \lambda k \left(o_r + g\right) \label{eq:removal}\\
  h_{t+2n} - h_{t}  &=& \lambda k \left(o_r + g - o_r + g \right)
                 = 2 \lambda kg
  \label{eq:offset_gone}
\end{eqnarray}
Thus the offset is removed by the chopper sign flips and the gradient
correctly recovered (\eqref{offset_gone}). However, this removal
becomes problematic in the step \eqref{removal} if $|o_r|\gg |g|$ and
$ \lambda n|o_r| \approx 1$ (1 being the threshold, see
\algref{TTiii}), since the update threshold would be falsely triggered
by the addition of the offset. Thus, for large offsets the learning
rate $\lambda$ needs to be decreased to avoid accidentally triggering
an update onto $\analog{w}$, which however slows down learning.

To improve upon this issue, we suggest an on-the-fly estimation of the
reference, where we use additional digital compute to store the value
of $\analog{a}$ right before the chopper sign flip and use this
value\footnote{or, in general, a leaky average of the recent past of
  $\analog{a}$} as a reference instead of the reference matrix
$\analog{R}$. The reasoning here is that the chopper flips are
unrelated to the direction of gradient information. Thus, if a
significant average gradient is currently present, the direction of
updates onto $\analog{A}$ has to change its direction when the chopper
flips. Thus the recent value of $\analog{a}$ can serve as good
reference point to read the information on $\analog{a}$ without bias.
This new algorithm is called \TTivfull. See \algref{TTiv} and
\figref{algillu} for implementation details.

\subsection{Learning rate scaling laws}

In the original formulation of the \TTii algorithm~\cite{gokmen2021},
the learning rate $\lambda_H$ for writing onto the hidden matrix $H$
was not specified explicitly. We here suggest to use the scaling law
\begin{equation}
  \label{eq:autogran}
  \lambda_H = \frac{\lambda\, \every\, n}{\autogran \weightgran}
\end{equation}
where $n$ is the number of columns of the weight matrix, $\every$ the
number of gradient updates done before a single column-read of
$\analog{A}$, and $\weightgran$ is the average update response size at
the \ac{SP} of $\analog{A}$.  Note that with \eqref{autogran} the
threshold of $H$ (ie. when to write onto $\analog{W}$) effectively
scales with how often a single element of the gradient matrix
$\analog{A}$ was updated before a read happens and is naturally
divided by a measure of the average update size $\weightgran$. Here
$\lambda$ is the learning rate of the standard \ac{SGD}, which might
be scheduled. We thus do not adjust the learning rate on $\analog{A}$
but instead the writing onto $H$ by the overall \ac{SGD} learning
rate. The hyper-parameter $\autogran$ specifies the length of
accumulation, with larger values averaging the read gradients for
longer. Note, however, that the same effect is done by adjusting
$\lambda$ so that tuning one of both is enough in practice.

Similarly, the learning rate $\eta$, used for the gradient update onto
$\analog{A}$, is important to set correctly for the \ac{DNN} at
hand. Since the pulse width is finite, the gradient magnitude has to
be large enough to cause a significant change on the gradient matrix
$\analog{A}$, so the learning rate $\eta$ needs to be scaled
accordingly. Since the gradient magnitude often differs for individual
layers, and might also change over time, we dynamically divide $\eta$
by the recent running average $\mu_x$ and $\mu_d$ of the absolute
maximum of the inputs $m_x = \max_j |x_j|$ and input gradients
$m_d = \max_i |d_i|$, respectively.
\begin{equation}
  \label{eq:eta}
  \eta = \frac{\eta_0 \lmax \weightgran}{\mu_x\mu_d}
\end{equation}
Note that $m_x$ and $m_d$ are needed for the gradient update already
(see~\algref{stocgradient}), so that this does not require any
additional computations, except for the leaky average
computations. Since $\lmax \weightgran$ is approximately the maximum
that the device material can change during one update ($\lmax$ is the
number of pulses used, see \algref{stocgradient}), the \eqref{eta}
means in case of $\eta_0=1$ that the a gradient update product
$x_id_j > \mu_x\mu_d$ is going to be clipped. The default value of
$\eta_0$ is 1, although is some cases higher values improve
learning. We use both of these scaling laws for all algorithms
presented here.

\section{Results}
\label{sec:results}
In following we present first a number of numerical test cases to
illustrate and compare the mechanism of the training algorithms. Then
we apply and compare them to SGD and DNN training with different
material and reference offsets settings. For simulations, we used the
\pytorch-based~\cite{pytorch} open source toolkit\footnote{The \TTii\
  algorithm is already implemented in the toolkit, however, we here
  improved on its GPU implementation by an runtime speed-up of up to
  $40\times$ by using a fused CUDA-kernel design (instead of the
  existing ``chunked'' kernel, see \figref{cuda_comparison}) and
  implemented our new algorithms. We also implemented a custom-design
  new device model, which explicitly subtracts the SP (which was not
  available in the toolkit).}  \AIHWKit~\cite{Rasch2021AFA}.

\subsection{Constant average gradient without offset}
\begin{figure}[t]
  \centering
  \includegraphics[width=0.5\textwidth]{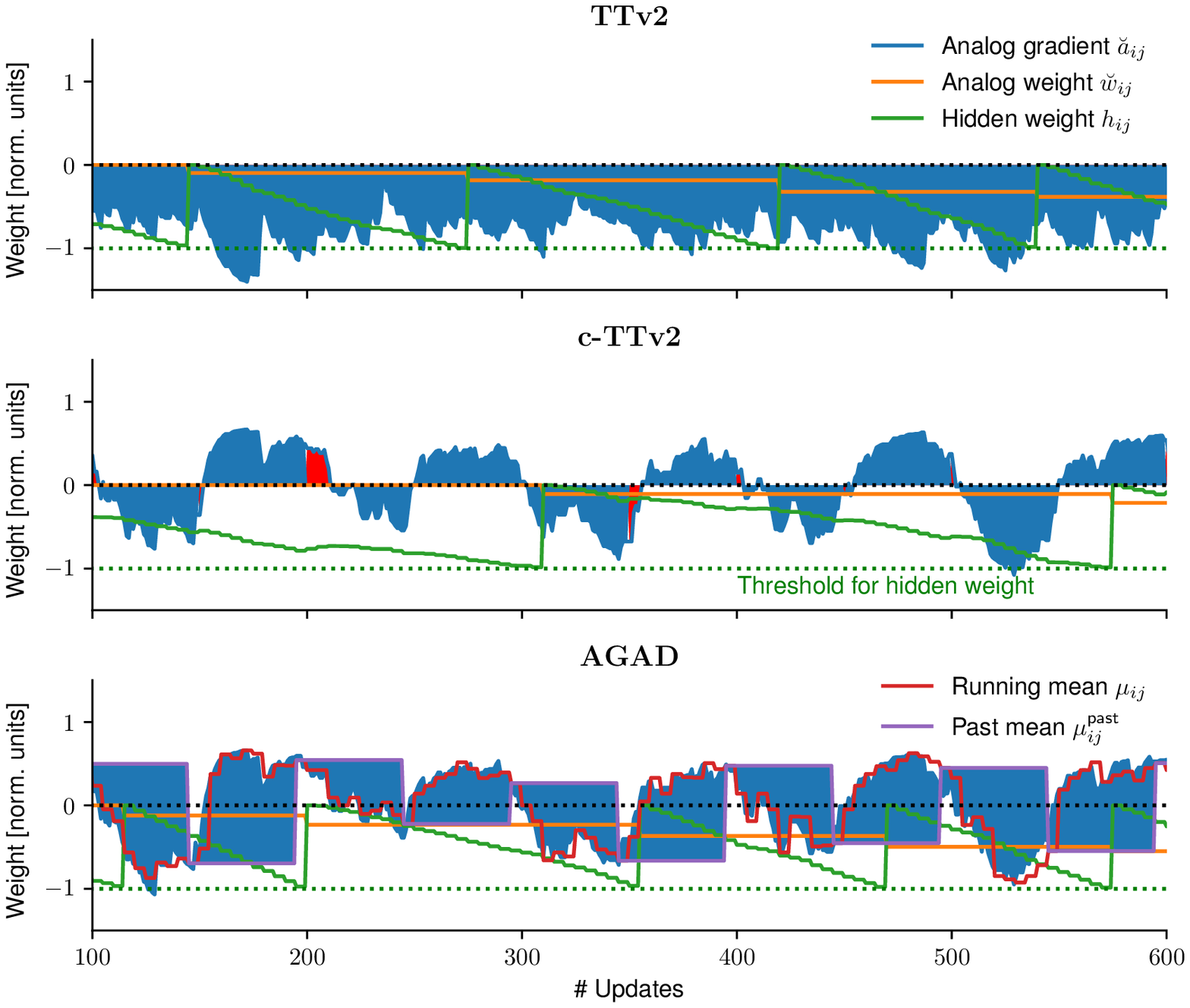}
  \vspace{-0.7cm}
  \caption{Mechanism of the algorithms with constant (negative)
    gradient input with perfect \ac{SP} estimation. Parameters:
    $\dwmin=0.05$, $\bdtod=\updowndtod=\dwmindtod=\dwminstd=0.3$, $\autogran=200$,
    $\lambda=0.1$, $\every=1$, $\leakylambda=0.5$, $\inchopprob=0.1$,
    $\lmax=5$, $\eta=1$, and $\mu_r = \sigma_r=0$.}
  \label{fig:corr_detect_nobias}
\end{figure}

As an illustration of the different learning algorithm we first
investigate a simple case where activations are given by $x = -X$ and
gradient inputs by $d = \alpha X + (1 - \alpha) Y$ where
$X, Y \sim {\cal N}(0, 1)$ are Gaussian random variables. Thus, in
this case the correlation of activations and gradients is given by
$\alpha$ and expected average update is only in one direction
$\Delta\analog{w} \propto -\alpha$. The various variables are plotted
against the number of updates for the three algorithms, \TTii, \TTiii\
and \TTiv\ in \figref{corr_detect_nobias}. Here we assume that the
reference matrix $\analog{R}$ is perfectly accurately set to the
\ac{SP} of $\analog{A}$ so that no offset remains.

Note that for \TTii\ (top panel of \figref{corr_detect_nobias}) the
trace of a selected $\analog{a}$ is strongly biased to negative
values, thus accumulating correctly the net gradient. It, however,
saturates at a certain level, caused by the characteristics of the
underlying device model (see \eqref{softbounds}). Because of the
occasional reads, the hidden weight accumulates until threshold is
reached at $-1$ (green trace), in which case the weight $\analog{w}$
is updated by one pulse (orange trace). The shaded blue area indicates
the value of $\omega = c \left(\analog{a} - \analog{r}\right)$ (where
the chopper $c=1$ is not used for \TTii). The area would be red if the
value was positive, which would indicate a hidden weight update in the
wrong direction.

In the middle panel of \figref{corr_detect_nobias}, the behavior of
the \TTiii\ algorithm is shown for the same inputs. Here, for better
illustration, a fixed chopper period is chosen. Since the incoming
gradient is constant (negative), the modulation with the chopper sign
causes an oscillation in the accumulation of the gradient in
$\analog{a}$. However, since the sign is corrected, the hidden matrix
is updated (mostly) into the correct direction (blue areas are sign
corrected). As we will see below, this flipping of signs will cancel
out offsets (here assumed to be 0). Due to transients in this example,
the trace of $\analog{a}$ has not always returned to the \ac{SP}
before reads, which causes some transient updates of the hidden
weights in the wrong direction (red areas). The weight $\analog{w}$ is
nevertheless correctly updated as the hidden weight averages out the
transients successfully. The rate of change of $\analog{w}$, however,
is somewhat impacted by the averaging of the transients.

Since \TTiv\ (\figref{corr_detect_nobias} bottom panel) uses an
(average) value $\mu^\text{past}$ of the recent accumulated gradient
$\analog{a}$ as the reference point (and not the static \ac{SP}
programmed onto $\analog{R}$) it is not plagued with the same
transients. In fact, the increased range causes a faster
update of the hidden matrix and subsequently the weight $\analog{w}$.

\begin{figure}[t]
  \centering
  \includegraphics[width=0.5\textwidth]{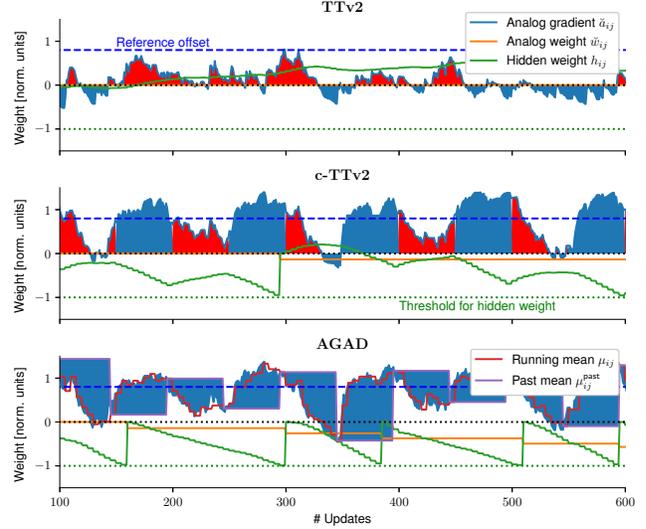}
  \vspace{-0.7cm}
  \caption{Mechanism of the algorithms with constant gradient input
    and significant reference offset. Parameters as in
    \figref{corr_detect_nobias} except $\mu_r =0.8$.}
  \label{fig:corr_detect_bias}
\end{figure}

\subsection{Constant average gradient with offset}

Since in \figref{corr_detect_nobias} the reference $\analog{R}$ was
set exactly to the \ac{SP}, no offset from the fixed-point
existed. In this case \TTii\ indeed works perfectly fine and might be
the algorithm of choice, because it requires least amount of digital
compute. However, now we examine the situation with a large offset,
so that the effective \ac{SP} of $\analog{A} - \analog{R}$ is
not at zero. In \figref{corr_detect_bias} we repeat the experiment of
\figref{corr_detect_nobias} with an offset of $\mu_r=0.8$ (dashed line
in \figref{corr_detect_bias}). We note that for \TTii, the constant
gradient pushes the accumulated gradient $\analog{a}$ away from the
\ac{SP} (blue dashed line), however, since the algorithm
does not take the offset into account, the update onto the hidden
matrix is wrong. In fact, hidden weight $h$ (green line) is even
updated in the wrong direction in this example (note that it is
expected to become more negative).

On the other hand, because of the effect of the chopper, even this
large bias can be successfully removed with the \TTiii\  algorithm
(\figref{corr_detect_bias} middle panel). Note that the hidden value
$h$ as well as the weight $\analog{w}$ decrease correctly. However,
due to the large bias, noticeable oscillations (red areas) are
stressing the accumulation on $h$, thus reducing the speed and
fidelity of the gradient accumulation. 

Due to the dynamic reference point computation of the \TTiv\
algorithm, the reference offset is almost irrelevant in this case
(\figref{corr_detect_bias} bottom panel), thus perfectly correcting
for any offset.

\subsection{Input-induced decay}
One hallmark of the \TTii\ algorithm is the input-induced decay:
random input or gradient fluctuations will draw the value of
$\analog{A}$ towards the \ac{SP} (see \eqref{spdecay}) and thus decay
to zero algorithmically (because of the reference $\analog{R}$, see
\eqref{reference}). This is shown in \figref{decay_high}, where it is
shown the hidden matrix decays quickly after gradients are suddenly
turned off. This is due to the noisiness of the input in this example
that causes $\analog{A}$ to decay quickly towards its \ac{SP}.

\begin{figure}[t]
  \centering
  \includegraphics[width=0.5\textwidth]{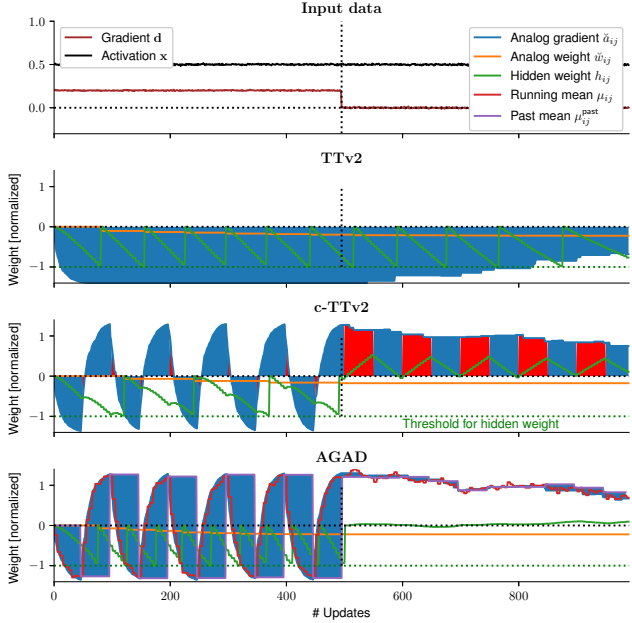}
  \vspace{-0.7cm}
  \caption{Sudden changes of gradient inputs. A long decay in the
    gradient matrix $\analog{A}$ for \TTii\ is caused if input
    fluctuations are low, possibly biasing the weight update in
    situations of sparse gradient inputs (note that the hidden weight
    threshold for \TTii\ is still hit after turning off net gradients
    at 500; green curve). Our new algorithms correct for this
    behavior. Parameters: as in \figref{corr_detect_nobias}. }
  \label{fig:decay_low}
\end{figure}

While activations and gradients generally are well approximated with a
noise process~(see discussion in \cite{onen2022neural}), in some
situations, however, gradients might be very sparse during \ac{DNN}
training, e.g. for CNNs with ReLU. If after a period of inputs,
suddenly no inputs were given to \TTii, for instance element $(i, j)$
of $dW$ is zero, the $\analog{a}_{ij}$ would not change, but would
indeed still continuously be read onto $h_{ij}$ which in turn would
wrongly update $\analog{w}_{ij}$. Such an (constructed) example is
shown in \figref{decay_low}, where the gradients are set to zero after
500 updates as before (\figref{decay_high}), but the noise of the
inputs is lowered. Note that, because of the input
fluctuation-dependent decay, the effective momentum-term of the \TTii\
algorithm becomes now very large (blue area after turning off the
gradients).While the update on the weight will eventually feed back to
the incoming gradients and correct for it, the quality of the gradient
update might nevertheless be impacted.

In our new algorithms (\TTiii\ and \TTiv), this artifact is not
present, since the gradient update onto the hidden matrix is modulated
by the chopper sign flips in both cases (\figref{decay_high} and
\figref{decay_low} lower panels). In particular, sparse gradients do
not cause any residual weight update, because if $\analog{a}_{ij}$ is
constant over multiple chopper periods it is treated as a bias that is
subtracted out. The subtraction happens on the hidden matrix in case
of \TTiii\ (notice the red and blue regions \figref{decay_low}). For
\TTiv\, a constant $\analog{a}_{ij}$ would not even be added to the
hidden matrix because the dynamic reference point adjustment when
changing the chopper phases removes any residual bias on-the-fly
(\figref{decay_low} bottom).

\subsection{Stochastic gradient descent on single linear layer}
\begin{figure}[t]
  \centering
  \includegraphics[trim=0.9cm 0 0 0, clip, width=0.5\textwidth]{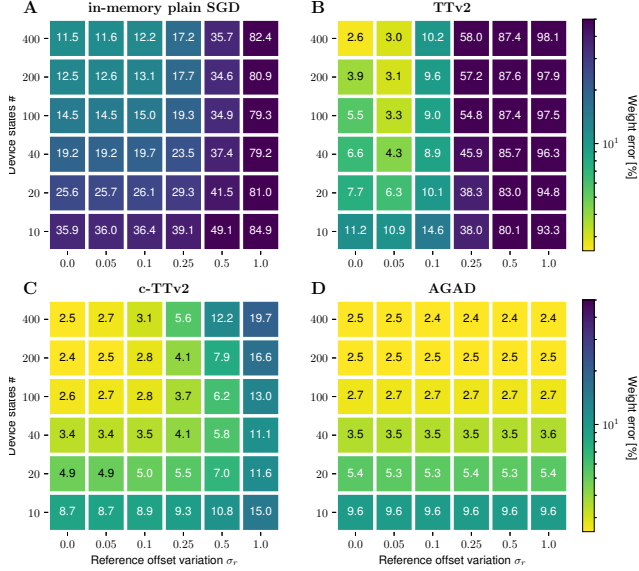}
  \vspace{-0.7cm}
  \caption{Weight programming error using different learning
    algorithms. Parameters: as in \figref{corr_detect_nobias} except
    that $\sigma_r$ and $\dwmin$ is varied. Additionally we set
    $\bdtod=0$ for $\analog{W}$ only (to not confound results for not
    being able to represent the target weight with $\analog{W}$). The
    target matrix, device-to-device variations, and inputs are fixed
    for each case for better comparison. Averaged over 3 runs.}
  \label{fig:weight_prog}
\end{figure}

Next we test how the algorithm perform when using \ac{SGD}. We consider
training to program a linear layer with output
$f_i(\mathbf{x}) = \sum_{j=1}^n w_{ij}x_j$ to a given target weight
matrix $\hat{W}$. We define the error as (mean squared deviation from
the expected output by using the target weight $\hat{W}$)
\begin{equation}
  \label{eq:loss}
   L(\mathbf{x}|W, \hat{W}) = \frac{1}{2m}\sum_i^m \Big(f_i(\mathbf{x}) - \sum_{j=1}^n\hat{w}_{ij}x_j\Big)^2
\end{equation}
Naturally, when minimizing the deviation (using \ac{SGD}) and updating
$W$, the error is minimized for $W=\hat{W}$ (assuming a full rank
weight matrix). We set $\hat{W}$ to random values ${\cal N}(0, 0.3)$
and use $x_j\sim {\cal N}(0, 1)$ as inputs. We evaluate the different
algorithms after convergence by the achieved weight error
$\weighterror^2 = \left\langle \left(w_{ij} -
    \hat{w}_{ij}\right)^2\right\rangle$, that is the standard
deviation of the learned weights with the target weight.
\figref{weight_prog} shows the results for a $20 \times 20$ weight
matrix.

We compared the three algorithms (\TTii, \TTiii, and \TTiv) as well as
plain in-memory \ac{SGD}, where the gradient update is directly done
on the weight $\analog{W}$ (\algref{sgd}). We explored the resilience
to two parameter variations, (1) the standard deviation of the
reference offset $\sigma_r$, as well as the number of device states
$\nstates$. As expected in case of no offset $\sigma_r=0$ and in
agreement with the original study~\cite{gokmen2021}, the \TTii\
algorithm works very well, vastly out performing in-memory \ac{SGD},
in particular for small number of states (e.g.
$\weighterror\approx 8\%$ vs $>25.0\%$, respectively, for 20 states
and the very same target weight matrix; see
\figref{weight_prog}AB). However, reference offset variations
$\sigma_r > 0$ critically affect the performance of \TTii. As soon as
$\sigma_r=0.1$, weight errors increase significantly, posing
challenges to current device materials and reference weight
programming techniques.

Using choppers, on the other hand, improves the resiliency to offsets
dramatically (\figref{weight_prog}CD). The \TTiii\ algorithm maintains
the same weight error for a large offsets in particular when the
number of states is small. Offsets in case of larger number of states
are less well corrected, consistent with the problems of transients
that increase for higher number of states~(see \eqref{spdecay}).  In
case of \TTiv, reference offset simply does not matter, as the
reference is dynamically computed on-the-fly. Moreover, in contrast to
\TTiii, \TTiv\ works equally well for higher number of states showing
that transients are not problematic either. There is nevertheless a
slight weight error increase for small number of states and small
offsets in comparison to \TTiii\ which is likely due to some
inaccuracies introduced by the on-the-fly estimation of the reference
value.

\subsection{\ac{DNN} training experiments}
\begin{figure}[t]
  \centering
  \includegraphics[width=0.5\textwidth]{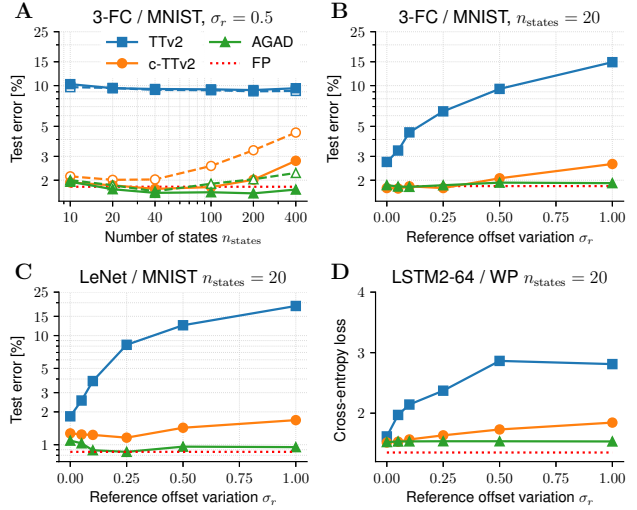}
  \vspace{-0.7cm}
  \caption{\ac{DNN} training with different analog learning
    algorithms.  SP of $\analog{W}$ is either corrected for (closed
    symbols) or not (open symbols with dashed lines). (A) Number of
    states versus test error with large reference offset variation
    $\sigma_r=0.5$. (B, C, D) Reference offset variation versus test
    error with $\nstates=20$. \acp{DNN}: (A, B) 3-layer
    fully-connected \ac{DNN} on MNIST (C) LeNet on MNIST (D) 2 layer
    LSTM on the War \& Peace dataset. }
  \label{fig:dnn_training}
\end{figure}

Finally, we compared the different learning algorithms for actual
\ac{DNN} training. For better comparison, we use largely the same
\acp{DNN} that was previously used to evaluate the earlier
algorithms. These were a fully-connected \ac{DNN} on
MNIST~\cite{gokmen2016}, LeNet on MNIST~\cite{gokmen2017cnn}, and a
two layer LSTM~\cite{gokmen2018lstm, gokmen2021}. We again trained the
\acp{DNN} with difference reference offsets variations~(see
\figref{dnn_training}; see \secref{sim-details} for details on the
simulations) with the same challenging device model (see example
device response traces for $\nstates=20$ in \figref{traces}). The
results are very consistent across the different \acp{DNN} and confirm
the trends found in case of the weight programming of one
layer~(compare to \figref{weight_prog}): If the offsets are perfectly
correct for, all algorithms fare very similarly. However, as expected,
the impact of a reference offset is quite dramatic for \TTii\, whereas
\TTiii\ can largely correct for it until it becomes too large. On the
other hand, \TTiv\ is not affected by the offsets at all and typically
shows best performance (\figref{dnn_training}B-D).

We found that even without offsets both new algorithms outperform the
state-of-the-art \TTii. Moreover, now that the gradients are computed
so well in spite of offsets and transients on $\analog{A}$, the second
order effect of the not corrected \ac{SP} on $\analog{W}$ is becoming
prevalent. Indeed, the test error improves beyond the \ac{FP} test
error for both \TTiii\ and \TTiv\ when the SP of $\analog{W}$ is
subtracted and thus corrected for (closed symbols), but increases
somewhat if not (open symbols). \TTiv\ shows better performance over
\TTiii\ in particular for larger number of states
(\figref{dnn_training}~A).

\section{Discussion}

We have introduced two new learning algorithms for parallel in-memory
training for crossbar arrays. All of these algorithms feature an
essential\footnote{We here assume for simplicity that the full matrix
  fits onto one crossbar.} ${\cal O}(n)$ time for calculating all
operations required for the full \ac{SGD}, namely forward, backward
and update. In particular, the ${\cal O}(n^2)$ calculation typically
needed for the outer product during the weight matrix update is
performed in ${\cal O}(1)$ since coincidence of a fixed number of
pulses is used. The read of one column is fully parallel again
${\cal O}(1)$, while the additional digital compute operates on one
column at a time and thus ${\cal O}(n)$.

Since \TTiii\ only introduces a sign flip, the computational
complexity added to \TTii\ is negligible. For \TTiv\ the reference
point is estimated on the fly which requires additional two digital
weight matrices, their loading and storing thus add to the
runtime. But since the additional operations still are performed on a
column at a time, \TTiv\ retains the ${\cal O}(n)$ complexity. One way
to reduce the digital compute of \TTiv, is to set $\leakylambda=1$ and
thus not compute any average over the recent past for the reference,
which reduces the additional storage requirements to one matrix (only
$\mu^\text{past}$), however, potentially introducing more noise.

The advantage of \TTiv\ is that it improves performance for higher
material states and makes the additional reference crossbar
$\analog{R}$ obsolete, potentially reducing the unit cell complexity.
Or alternatively to removing $\analog{R}$, this crossbar could now be
used to store the SP of $\analog{W}$ and thus further improving
accuracy with the same unit cell design.

We find that both \TTiii\ and \TTiv\ push the boundary of in-memory
training performance, while relaxing the number of state requirements
on device materials.  Indeed, we show that 20 states are now
sufficient for training almost to floating point accuracy for our
example \acp{DNN}, even in the presence of \ac{SP} fluctuations or
long-term transients.

\bibliography{ttv34}
\bibliographystyle{mlsys2023}

\newpage
\appendix
\renewcommand\thefigure{\thesection.\arabic{figure}}
\section{Appendix}
\setcounter{figure}{0}

\subsection{\ac{DNN} training simulation details}
\label{sec:sim-details}
Here we describe the details of the \ac{DNN} training simulations done in \figref{dnn_training}.

\subsubsection{3-FC / MNIST}
\label{sec:dim-details-fc}
Here we use a 3-layer fully-connected \ac{DNN} on the MNIST data set
with sigmoid activations and hidden sizes of 255 and 127 as described
in \cite{gokmen2016}. For the analog \ac{MVM} (forward and backward
pass), we use essentially the standard settings of \AIHWKit, which
includes output noise (0.5 \% of the quantization bin width),
quantization and clipping (output range set to 20, output noise to
0.1, and input and output quantization to 8 bit). It uses the noise
and bound management techniques as described in~\cite{gokmen2017cnn}.
To adjust the output range to the \ac{DNN} range, we add a learnable
(scalar) floating point factor after each crossbar output, which is
updated with standard momentum \ac{SGD} (0.9 momentum term).

We train for 80 epochs and schedule the learning rate into 3 steps
(35, 35, and 10 epochs), with reduction factor of 0.1. Starting
learning rate is set to 0.05 (0.1 for floating point) and batch size
is 10. Additionally, we set $\every=1$ and $\autogran=10000$ and
confirmed by a test run that it was a reasonable setting.  We then run
a grid of simulations and report the average test accuracy over the
last 3 training epochs. The grid was the same for each condition,
namely we simulated the settings of
$\eta=0.05, 0.1, 0.2, 0.5, 1.0, 2.0$. For each condition the best test
error (averaged averaged over the last 3 epochs) is reported in the
figure. The number of states was varied as indicated in the figure
(see \figref{dnn_training}; parameter $\weightgran$).  Other
parameters are set as reported in \figref{corr_detect_nobias}. The
chopper probability was set to $\inchopprob=0.1$, where random
switching was used for \TTiii\ and regular switching for \TTiv\ as
described in the algorithms (see \algref{TTiii} and \algref{TTiv},
respectively).

For \figref{dnn_training}~A, we additionally varied the learning rate
for each condition, and took the better one of the two
$\lambda=0.05, 0.025$. This resulted thus in 432 training simulations
for this subplot alone.

\subsubsection{LeNet / MNIST}
A variant of the LeNet5~\cite{lecun1998gradient} model architecture,
which contains 2-convolutional layers, 2-max-pooling layers, and
2-fully connected layers, trained on the MNIST dataset, is used
here. The original LeNet5 model has 3-convolutional layers in its
architecture. The model training is performed using the same setting
of \AIHWKit\ for the analog \ac{MVM} (forward and backward pass) and
the same device model as described above (\secref{dim-details-fc}).

The model is trained for 60 epochs using a batch size of 8 and
$\autogran=10000$. Different learning rate $\lambda$ values of 0.01,
0.02, 0.03, 0.04, and 0.05 are explored for $\sigma_r=0.0$, and the
learning rate with the best test accuracy value is picked for each
method under consideration, which resulted in the values $0.03$,
$0.04$, and $0.04$ for \TTii, \TTiii, and \TTiv, respectively. The
learning rate is scheduled into two steps (45 and 15 epochs) with a
reduction factor of 0.10.  The chopper probability (for the \TTiii\
and \TTiv\ cases) is set to $\inchopprob=0.1$, with random switching
used for \TTiii\ and regular switching for \TTiv.

The initial values of $\eta$ and $\every$ are set to 1, and 1,
respectively. Firstly, the value of $\eta$ is tuned until the best
result is obtained. After that, the value of $\every$ is also
tuned. This tuning sequence is encouraged as $\eta$ has more influence
on the model performance as measured using the test accuracy than
$\every$. The combination of these hyperparameter values that gives
the best result is selected for all the methods. We find that
$\every=1$ works best for all methods and set $\eta$ to $0.05$, $0.1$
and $0.075$ for \TTii, \TTiii, and \TTiv, respectively. The test
accuracy reported is the average over the test errors obtained after
the last three training epochs.

The above obtained values of the $\eta$ and $\every$ hyperparameters
are then used to obtain the test accuracy for
$\sigma_r=0.05, 0.1, 0.25, 0.5, 1.0$ respectively, and the resulting
test accuracy is used to generate the plot in \figref{dnn_training}~C.

\subsubsection{LSTM / War \& Peace}
We use a LSTM network composed of 2 stacked LSTM blocks with hidden
vector size of 64 followed by a fully-connected layer, as described
in~\cite{gokmen2018lstm}. We trained this network on the War and Peace
(WP) novel. The dataset is split into a training set and test set with
2,933,246 and 325,000 character respectively, and a total vocabulary
of 87 characters.

We trained the network using the \AIHWKit\ for 100 epochs and schedule
the learning rate into 3 steps (50, 40, and 20 epochs) with a
reduction factor of 0.1. The starting learning rate is set to 0.1,
batch size to 16, sequence length to 100, $\autogran=10000$ and
chopper probability (for the \TTiii\ and \TTiv\ cases) is set to
$\inchopprob=0.1$, with random switching used for \TTiii\ and regular
switching for \TTiv.

We run multiple simulations starting from the condition with reference
offset variation $\sigma_r=0$. We explore different learning rate
$\lambda=0.05, 0.1, 0.75$, different settings for $\every=1, 3$ and
for $\eta=0.05, 0.1, 0.5, 1.0$ and take the best combination of
parameter for each algorithm. With the found parameter, we then
perform the simulations for various reference offset variations and
take the loss of the last epoch to generate the plot in
\figref{dnn_training}~D.

\subsection{Supplementary figures}
\begin{figure}[ht]
  \centering
  \includegraphics[width=0.5\textwidth]{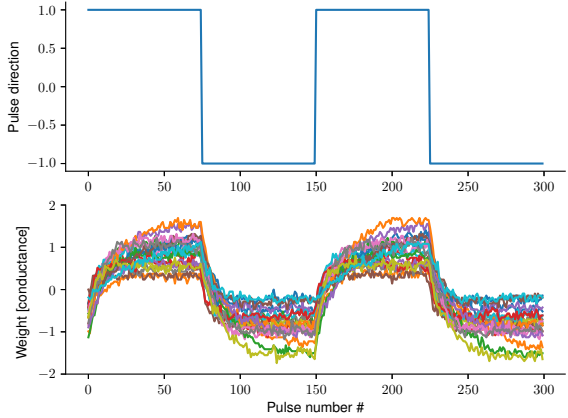}
  \vspace{-0.7cm}
  \caption{Response traces of the simulated material used
    ($\nstates=20$). Upper plot shows the pulsing
    pattern. Lower plot shows the response of 20 devices
    (colors). Note that significant asymmetry, device-to-device variation and
    cycle-to-cycle variation is present.  }
  \label{fig:traces}
\end{figure}

\begin{figure}[ht]
  \includegraphics[width=0.5\textwidth]{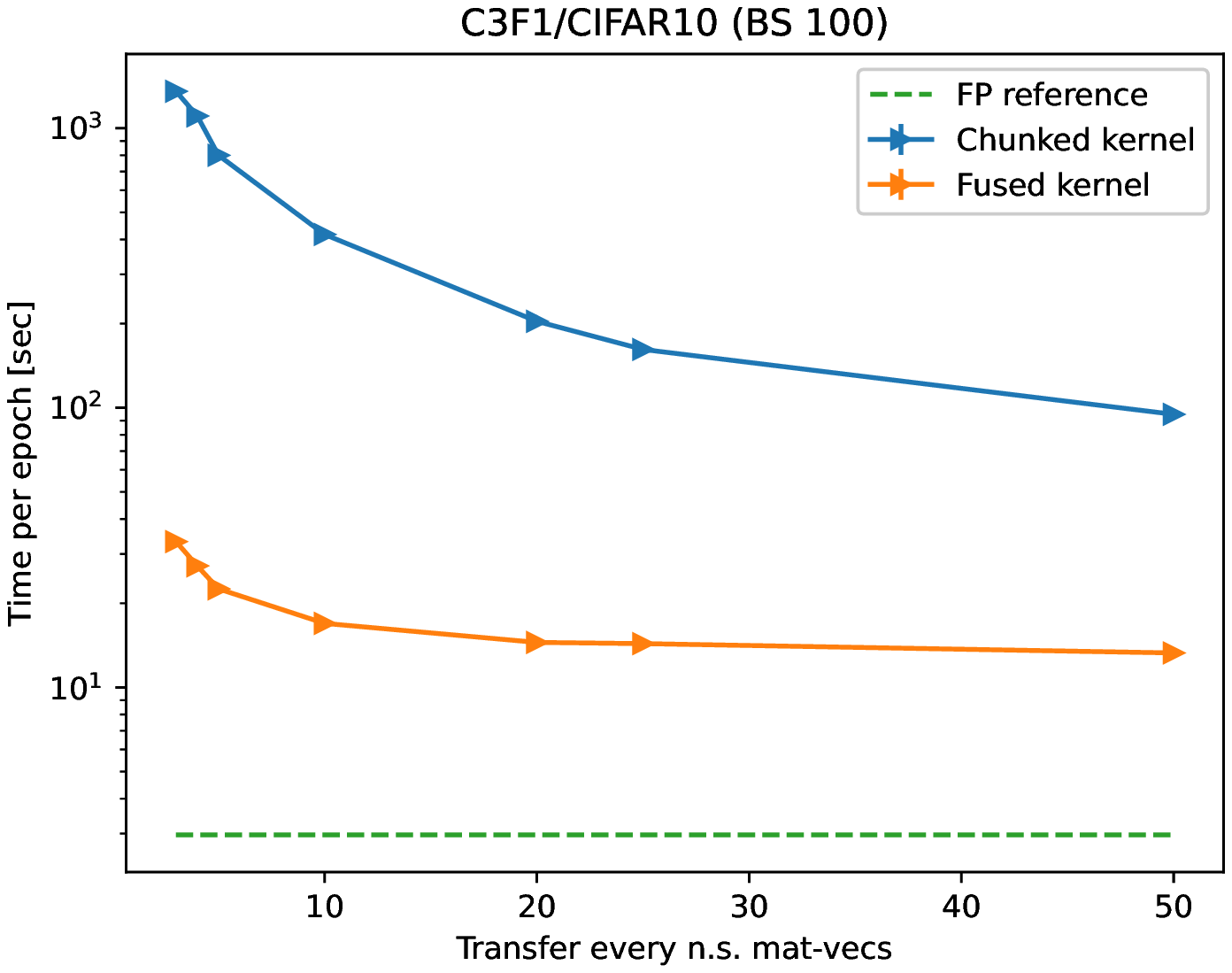}
  \vspace{-0.7cm}
  \caption{Runtime of our improved CUDA kernel design for \TTii\
    (``Fused kernel'') in comparison to the existing ``Chunked
    kernel'' of the open source toolkit \AIHWKit. The fused kernel is
    also used for our new algorithms \TTiii\ and \TTiv. Here the time
    for one epoch on a standard small CNN on the CIFAR10
    dataset~\cite{krizhevsky2009learning} is shown. The runtime is
    evaluated for different $\every$ parameters (compare to
    \figref{algillu}). Note that our fused design improves the speed
    dramatically, from $>1000$ seconds to only $30$ seconds in the
    best case. }
  \label{fig:cuda_comparison}
\end{figure}

\begin{figure}[ht]
  \centering
  \includegraphics[width=0.5\textwidth]{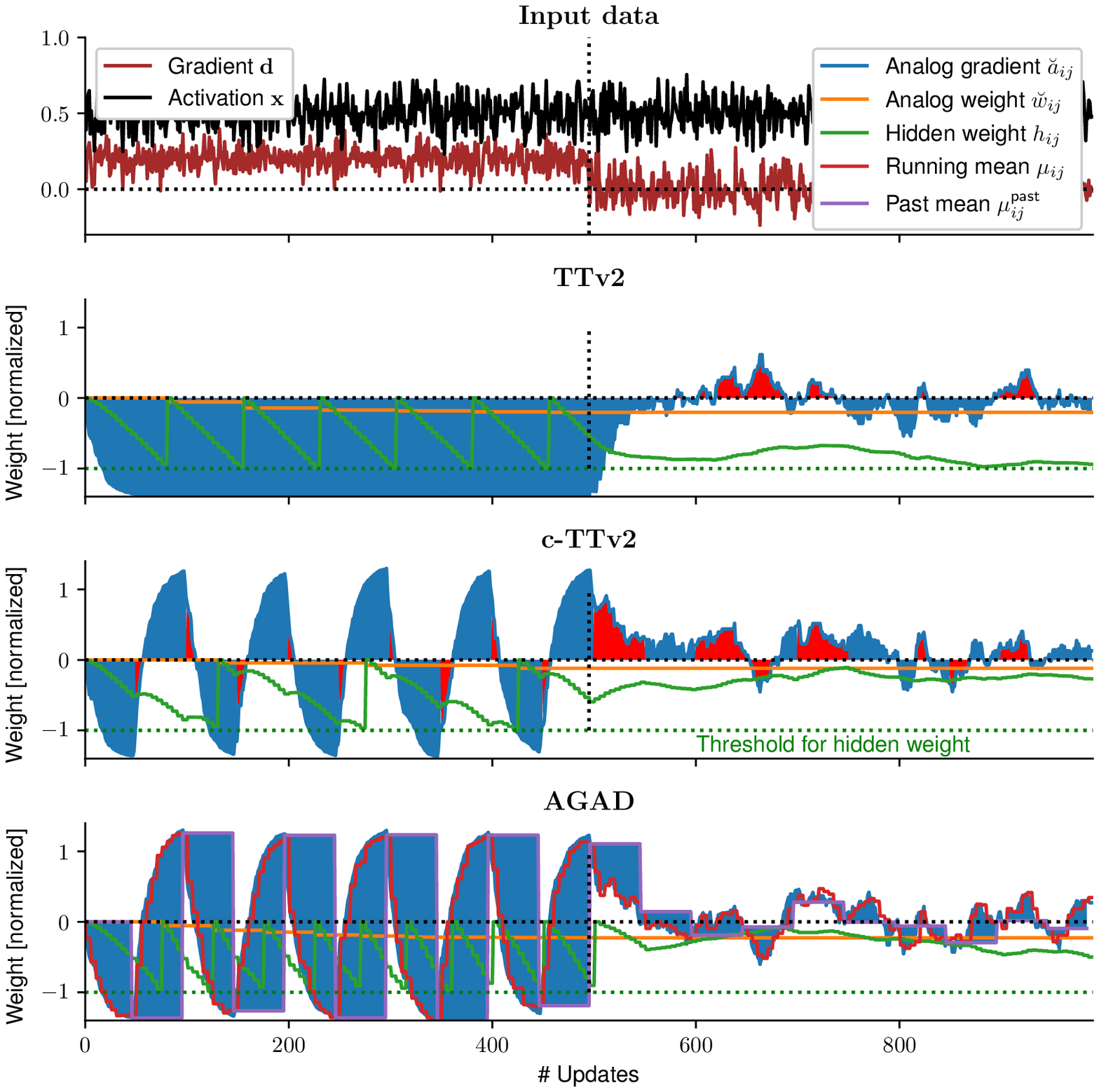}
  \vspace{-0.7cm}
  \caption{Switching gradient input off causes the gradient
    matrix $\analog{A}$ to decay for \TTii\ if input fluctuations are
    high. Our new algorithms also correctly stop the weight update
    in this case. Parameters: as in \figref{corr_detect_nobias}.}
  \label{fig:decay_high}
\end{figure}


\end{document}